\pdfoutput=1

\documentclass[11pt]{article}

\usepackage[preprint]{acl}

\usepackage{times}
\usepackage{latexsym}
\usepackage{threeparttable}
\usepackage[T1]{fontenc}

\usepackage[utf8]{inputenc}

\usepackage{microtype}

\usepackage{inconsolata}

\usepackage{graphicx}
\usepackage{times}
\usepackage{latexsym}
\usepackage{multirow}
\usepackage{caption}
\usepackage{makecell}
\usepackage{graphicx}%
\usepackage{multirow}%
\usepackage{amsmath,amssymb,amsfonts}%
\usepackage{amsthm}%
\usepackage{mathrsfs}%
\usepackage[title]{appendix}%
\usepackage{xcolor}%
\usepackage{textcomp}%
\usepackage{manyfoot}%
\usepackage{booktabs}%
\usepackage{algorithm}%
\usepackage{algorithmicx}%
\usepackage{algpseudocode}%
\usepackage{listings}%
\usepackage{microtype}%
\usepackage{array}
\usepackage{lipsum}
\usepackage{longtable}
\usepackage{float}
\usepackage{adjustbox}
\usepackage[T5]{fontenc} 
\usepackage{mathptmx}
\usepackage{subfigure}
\usepackage{amsmath}
\usepackage{pifont}
\usepackage{tabularx}
\usepackage{arydshln}

\title{A study of Vietnamese readability assessing through semantic and statistical features}

\author{Hung Tuan Le$^{1,3,^\clubsuit}$, Long Truong To$^{1,3,^\clubsuit}$, Manh Trong Nguyen$^{1,3,^\clubsuit}$\\ \textbf{Quyen Nguyen$^{2,3,^\diamondsuit}$, Trong-Hop Do$^{1,3,^\spadesuit}$}\\
         $^1$University of Information Technology, Ho Chi Minh City, Vietnam\\
         $^2$International University, Ho Chi Minh City, Vietnam\\ 
         $^3$Vietnam National University, Ho Chi Minh City, Vietnam \\ 
         \{21521101, 21520250, 21520343\}@gm.uit.edu.vn$^\clubsuit$ \\
         ntquyen@hcmiu.edu.vn$^\diamondsuit$
         hopdt@uit.edu.vn$^\spadesuit$\\}

\begin{document}

\maketitle
\begin{abstract}
Determining the difficulty of a text involves assessing various textual features that may impact the reader's text comprehension, yet current research in Vietnamese has only focused on statistical features. This paper introduces a new approach that integrates statistical and semantic approaches to assessing text readability. Our research utilized three distinct datasets: the Vietnamese Text Readability Dataset (ViRead), OneStopEnglish, and RACE, with the latter two translated into Vietnamese. Advanced semantic analysis methods were employed for the semantic aspect using state-of-the-art language models such as PhoBERT, ViDeBERTa, and ViBERT. In addition, statistical methods were incorporated to extract syntactic and lexical features of the text. We conducted experiments using various machine learning models, including Support Vector Machine (SVM), Random Forest, and Extra Trees and evaluated their performance using accuracy and F1 score metrics. Our results indicate that a joint approach that combines semantic and statistical features significantly enhances the accuracy of readability classification compared to using each method in isolation. The current study emphasizes the importance of considering both statistical and semantic aspects for a more accurate assessment of text difficulty in Vietnamese. This contribution to the field provides insights into the adaptability of advanced language models in the context of Vietnamese text readability. It lays the groundwork for future research in this area.
\end{abstract}

\section{Introduction}

Exchanging information and knowledge through texts has led to the emergence of measuring text difficulty. There can be multiple ways to describe and convey content when dealing with the same issue. Among them, complex texts pose challenges for readers, as reflected in lower reading speed, poorer comprehension, and reduced capacity to connect information within the text. In recent years, text difficulty has been evaluated through linguistically motivated features, such as syntactic complexity, complexity in logical relationships and inferences of information in the text, and the sequential expression of data over time or context. Two main approaches for determining text difficulties have been proposed, namely statistical approach and machine learning or deep learning. In the former approach, text difficulty is evaluated through the synthesis of easy-to-compute features in the text, such as the length of the text, the average number of words and sentences in the text, etc. \cite{flesch1948new, kincaid1975derivation}, where these features are extracted and evaluated through correlation analysis with the difficulty of a set of texts. The second approach, namely machine or deep learning approach, involves using neural models to represent the semantics present in the text, allowing for the assessment of text difficulty \cite{heilman2007combining, heilman2008analysis, lee-etal-2021-pushing, si2001statistical}.

Studies addressing the problem by applying advanced neural models such as BERT and its variants combined with features extracted through traditional statistical methods have achieved promising results on English datasets such as WeeBit \cite{vajjala2012improving}, OneStopEnglish \cite{vajjala2018onestopenglish}, and Cambridge \cite{xia2016text}. In Vietnam, pioneering research in this area, such as that of \cite{nguyen1985second,luong2018new}, and more recently \cite{luong2022}, has applied PhoBERT, which is a pre-trained language model \cite{nguyen-tuan-nguyen-2020-phobert} designed specifically for Vietnamese, to address the problem. However, these studies assess text difficulty of sentences in isolation while overlooking features that span over an extended discourse, such as discourse relations or entity cohesion across a series of sentences.

Given the gap in previous literature on Vietnamese text readability assessment, this study scrutinizes the impacts of statistical and semantic features, as well as the correlation between these two types of features on the difficulty of Vietnamese texts across three primary datasets: Vietnamese Readability dataset \cite{luong2020building}, RACE \cite{lai2017race}, and OneStopEnglish \cite{vajjala2018onestopenglish}. Our methods range from traditional machine learning models such as SVM, Random Forest, and Extra Tree to state-of-the-art pre-trained language models in various semantic tasks, such as PhoBERT \cite{nguyen-tuan-nguyen-2020-phobert}, ViDeBERTa \cite{tran2023videberta}, and ViBERT \cite{tran2020improving}. The joint approach combining statistical and semantic features are shown to improve model performance, although not yet surpassing statistical features alone. However, they demonstrate potential for development on larger datasets.

Furthermore, we conduct an in-depth analysis of specific groups of statistical features concerning text difficulty by individually examining each feature group across multiple models. The results show that features such as 'Number of words' or 'Average word length in characters' have the most significant impact on the models when combined with semantic features from deep learning models.

\section{Related Works} 
\label{sec:Related Works}
This section provides an in-depth analysis of global body of research addressing the challenges of readability (see section \ref{subsec:Textual Readability}), with a particular focus on the existing study conducted within the Vietnamese context (see section \ref{subsec:Vietnamese Readability}).
\subsection{Textual Readability} 
\label{subsec:Textual Readability}
Research on textual readability has increasingly captured of scholars within the natural language processing domain. This interest is particularly evident in foundational English-language studies, such as those pioneered by \citeauthor{flesch1948new}, which adopted a statistical lens to investigate the problem. These early investigation focused on evaluating text readability by quantifying linguistic features such as syllable per word ratio. Later, in 1975, the readability index by \citeauthor{kincaid1975derivation} was published based on the features of \citeauthor{flesch1948new}. In \citeauthor{chall1995readability} (1995), the readability of the text was assessed based on the semantic difficulty of words in the text by examining the frequency of word occurrences with a word list of 3000 words. In the following years, these features became standards for evaluation \cite{fry1990readability,lennon2004lexile}, along with syntactic features such as the height of the parse tree \cite{chall1995readability}. However, the statistical approach remains limited in its ability to capture deeper linguistic features that critically influence text readability, such as discourse relations, cohesion, and rhetorical structure \cite{collins2014computational}.

As language models have advanced and training data volumes have expanded, a new approach to the readability problem has emerged. This approach harnesses the language representation capabilities of these models to extract deeper linguistic features while utilizing the classification power of probabilistic and deep learning models. Early studies include those by \citeauthor{si2001statistical} and \citeauthor{collins2004language} who applied unigram language models and classification through naive Bayes. In the following years, the probabilistic model approach gained attention and achieved good results \cite{schwarm2005reading,heilman2007combining, heilman2008analysis,pilan2014rule}. Since the rise of deep learning models, particularly with the advent of pre-trained language models utilizing transformer architecture, which have achieved state-of-the-art results across various semantic tasks, the performance on the readability problem has notably improved. This enhancement is due not only to the advanced feature extraction capabilities of these models \cite{cha2017language,jiang2018enriching,azpiazu2019multiattentive} but also to their integration with externally collected statistical features \cite{deutsch2020linguistic,meng2020readnet,lee-etal-2021-pushing}.

Beyond English, research has also expanded to other languages, building upon the established foundation of English-language studies, with notable developments in languages such as French \cite{franccois2012ai}, Italian \cite{dell2011read}, German \cite{hancke2012readability}, Swedish \cite{falkenjack2013features,pilan2016readable}, Bangla \cite{islam2012text}, and Greek \cite{chatzipanagiotidis2021broad}.

\subsection{Vietnamese Readability}
\label{subsec:Vietnamese Readability}
Research on the readability problem remains limited, primarily due to the scarcity of high-quality datasets. This issue is evident in studies ranging from \cite{nguyen1985second,nguyen1982readability} to \cite{luong2020building,luong2018new,nguyễn2019affection}, where dataset sizes have been notably small, often comprising fewer than 2,000 samples. Furthermore, the dominant approach to addressing the readability problem has centered on feature extraction through statistical analysis. This includes metrics such as the number of syllables or words, the height and width of parse trees, and the count of clauses \cite{luong2020assessing}. Recently, \citeauthor{luong2022} adopted a novel approach to the problem by extracting features using PhoBERT \cite{nguyen-tuan-nguyen-2020-phobert}. However, this research has yet to be made accessible to the broader community.
\section{Current Study}
\label{sec:Current Study}
In this section, we describe the experimental process in the paper, including the datasets (see section \ref{subsec:Datasets}) and the methods we experimented with (see section \ref{subsec:Empirical Method}).
\subsection{Datasets}
\label{subsec:Datasets}

We use a total of three datasets described in Table \ref{tab:Datasets statistics}, namely OneStopEnglish \cite{vajjala2018onestopenglish}, RACE \cite{lai2017race}, and the Vietnamese Text Readability Dataset \cite{luong2020building}.

The Vietnamese Text Readability Dataset (ViRead) \cite{luong2020building} is constructed from Vietnamese college-level textbooks, stories, and literature websites. After extracting text from these sources using OCR, a team of twenty Vietnamese literature teachers from middle schools, high schools, and colleges labels the sentences. The labels are categorized into four levels: Very Easy, Easy, Medium, and Difficult.

Due to the lack of large-scale and high-quality datasets in Vietnamese for the readability problem, we also use two English datasets: OneStopEnglish \cite{vajjala2018onestopenglish} and RACE \cite{lai2017race}. The OneStopEnglish dataset is extracted from onestopenglish\footnote{https://onestopenglish.com/}, an English language learning resources website run by MacMillan Education. The content has been rewritten into three versions from \textit{The Guardian} newspaper, each labeled as advanced (Adv), intermediate (Int), and elementary (Ele). The RACE dataset, a large-scale reading comprehension benchmark, is derived from English exams administered to Chinese middle and high school students and includes 28,000 passages. For the readability task, RACE is divided into junior and senior levels.

We translated the two English datasets, OneStopEnglish and RACE, into Vietnamese using Google Translate\footnote{https://translate.google.com/}. Subsequently, we partitioned these datasets into smaller components for the experimentation process. Given the limited size of the OneStopEnglish and ViRead datasets, each containing fewer than 2,000 samples, we divided them into two sets: a training set (train) and a test set (test). The size statistics for each dataset are provided in Table \ref{tab:Datasets statistics}.
\begin{table*}[htb]
\centering
\resizebox{\textwidth}{!}{%
\begin{tabular}{clccccc}
\hline
\textbf{Datasets} & \multicolumn{1}{c}{\textbf{Domain}} & \textbf{Language} & \textbf{Number of sample} & \textbf{Number of class} & \textbf{Training} & \textbf{Test} \\ \hline
ViREAD & Literature & Vietnamese & 1825 & 4 & 1460 & 365 \\
Race & Education & English & 27933 & 2 & 22346 & 5587 \\
OneStopEnglish & Educaion & English & 567 & 3 & 453 & 114 \\ \hline
\end{tabular}%
}
\caption{Datasets statistics}
\label{tab:Datasets statistics}
\end{table*}

\subsection{Empirical Method}
\label{subsec:Empirical Method}
In this section, we proceed to design the implementation process along two main approaches: the statistical approach (see section \ref{subsubsec:Statistical approach}) and the semantic approach (see section \ref{subsubsec:Semantic approach}). The statistical approach involves employing statistical methods to extract features from the dataset, whereas the semantic approach leverages machine learning models, ranging from basic to advanced deep learning techniques, to derive semantic features. Additionally, we conduct experiments that integrate features from both statistical and semantic approaches to examine their correlation and impact on the results (see section \ref{subsubsec:joint approach}).
\subsubsection{Statistical approach}
\label{subsubsec:Statistical approach}
\citeauthor{luong2020assessing} performed experiments to evaluate the impact of various features on text readability using a statistical approach, specifically on the Vietnamese readability dataset \cite{luong2020building}. The features examined included part-of-speech features (such as the ratio of POS-tagged words and the proportion of common nouns to distinct words), syntax-level features (including average parse tree depths), and Vietnamese-specific features (like the ratio of borrowed words and Sino-Vietnamese words). We selected features that exhibited a high correlation with text difficulty, as detailed in Table \ref{tab:Linguistic features}.

Additionally, we introduced two new features related to word cohesion, represented through dependency trees, to investigate how the relationships between words within a sentence impact text difficulty (see table \ref{tab:Linguistic features}). To extract these two features, we utilized VnCoreNLP \cite{vncorenlp} for sentence segmentation and dependency representation. The statistical features will be classified using three machine learning models: Support Vector Machine (SVM), Random Forest, and Extra Trees.

\begin{table*}[htb]
\centering

\resizebox{\textwidth}{!}{%
\begin{tabular}{cl}
\hline
\textbf{Category}                            & \multicolumn{1}{c}{\textbf{Feature}}                      \\ \hline
\multirow{3}{*}{Raw Feature}                 & Number of words                                           \\
                                             & Average word length in character                          \\
                                             & Ratio of long sentence (in syllable)                      \\ \hline
\multirow{4}{*}{POS Feature}                 & Distinct common nouns/distinct words                      \\
                                             & Distinct parallel conjunctions/distinct words             \\
                                             & Ratio of single POS tag words                             \\
                                             & Adverbs/sentences                                         \\ \hline
\multirow{2}{*}{Syntax-Level Feature}       & Average no. distinct conjunction word                     \\
                                             & Average no. conjunction word                              \\ \hline
\multirow{3}{*}{Vietnamese-Specific Feature} & Ratio of borrowed words                                   \\
                                             & Ratio of distinct borrowed words                          \\
                                             & Ratio of distinct Sino-Vietnamese words                   \\ \hline
\multirow{2}{*}{Word Cohension}              & Depth of Dependency Tree                                  \\
                                             & Average overlapping between multiple sentences in paragraph \\ \hline
\end{tabular}%
}
\caption{Linguistic features}
\label{tab:Linguistic features}
\end{table*}

The statistical features on the three datasets ViRead, OneStopEnglish, and RACE are summarized in Table \ref{tab:The extraction result of statistical features in ViRead, OneStopEnglish and RACE}. As noted, in translated datasets such as OneStopEnglish and RACE, some standard text features remain consistent, such as 'Average word length in characters' and 'Distinct parallel conjunctions/distinct words.' For the 'Ratio of long sentences' feature, we define sentences with more than 20 syllables, based on research from the American Press Institute. However, features specific to Vietnamese, such as the 'Ratio of borrowed words' and the 'Ratio of distinct Sino-Vietnamese words,' vary. This variation is attributed to translation nuances and unique characteristics of Vietnamese texts. These differences significantly impact the models' results, as discussed in Section \ref{sec:Experiment Result}.

\begin{table*}[htb]
\centering
\resizebox{\textwidth}{!}{%
\begin{tabular}{l|c|c|c}
\hline
\multicolumn{1}{c|}{\textbf{Feature}} & \textbf{ViRead} & \textbf{OneStopEnglish} & \textbf{RACE} \\ \hline
Number of words & 40 - 23104 & 263 - 1417 & 13 - 1271 \\ 
Average word length in character & 2.4973 - 3.4071 & 2.9754 - 3.501792 & 2.287 - 5.483 \\
Ratio of long sentence (in syllable) & 0 - 1 & 0.2714 - 1 & 0 - 1 \\ \hdashline
Distinct common nouns/distinct words & 0.0312 - 0.44 & 0.1194 - 0.2612 & 0 - 0.5 \\
Distinct parallel conjunctions/distinct words & 0- 0.1129 & 0.0052 - 0.0284 & 0 - 0.1739 \\
Ratio of single POS tag words & 0.7977 - 1 & 0.8815 - 0.9627 & 0.8421 - 1 \\ 
Adverbs/sentences & 1 - 82 & 7 - 34 & 0 - 39 \\ \hdashline
Average no. distinct conjunction word & 0 -36 & 3 -18 & 0 -18 \\
Average no. conjunction word & 0 -1670 & 11 - 77 & 0 - 79 \\ \hdashline
Ratio of borrowed words & 0 - 0.0128 & 0 - 0.0279 & 0 - 0.0058 \\
Ratio of distinct borrowed words & 0 - 0.0085 & 0 - 0.0085 & 0 - 0.044 \\
Ratio of distinct Sino-Vietnamese words & 0.0317 - 0.4179 & 0.0022 - 0.0149 & 0 - 0.396 \\ \hdashline
Depth of Dependency Tree & 1.5 - 30.3333 & 6.8966 - 21.1053 & 1 - 132 \\ 
Average overlapping between multiple setence in paragraph & 0.2539 - 143.2710 & 1.6590 - 10.5664 & 0 - 11.157 \\ \hline
\end{tabular}%
}
\caption{The min-max extraction result of statistical features in ViRead, OneStopEnglish and RACE}
\label{tab:The extraction result of statistical features in ViRead, OneStopEnglish and RACE}
\end{table*}
\subsubsection{Semantic approach}
\label{subsubsec:Semantic approach}
In this section, we employ advanced semantic analysis methods for classifying the difficulty level of Vietnamese texts. Our semantic approach primarily utilizes three state-of-the-art language models: PhoBERT \cite{nguyen-tuan-nguyen-2020-phobert}, ViDeBERTa \cite{tran2023videberta}, and ViBERT \cite{tran2020improving}. These models are instrumental in extracting deep semantic features from the Vietnamese texts, which are crucial for our classification task.

PhoBERT \cite{nguyen-tuan-nguyen-2020-phobert} emerges as a paragon, trained extensively on a corpus comprising 20GB of Vietnamese Wikipedia and news texts. It boasts 135 million parameters in its base iteration and an augmented 370 million parameters for the large variant. In its most recent iteration, PhoBERT$_{base}-V2$, the model has been refined on a formidable 120GB of Vietnamese texts derived from the OSCAR-2301 dataset\footnote{https://huggingface.co/datasets/oscar-corpus/OSCAR-2301}.

ViDeBERTa \cite{tran2023videberta} is a model with the architecture of DeBERTa \cite{he2020deberta} and has been trained on CC100\footnote{https://huggingface.co/datasets/cc100} corpus, including 138GB uncompressed texts. ViDeBERTa outperforms PhoBERT on tasks such as named entity recognition (NER) and part-of-speech (POS). However, the current version of ViDeBERTa with the DeBERTa-V3 architecture has not been released; instead, the version with the DeBERTa$_{Base}$-V2 architecture is available \footnote{https://huggingface.co/Fsoft-AIC/videberta-base}. ViBERT \cite{tran2020improving} has been trained on approximately 10GB of texts collected from online newspapers in Vietnamese, enabling the model to represent the semantics of words more effectively.

The features extracted from pre-trained language models will be classified using a range of machine learning models, including Support Vector Machine (SVM), Random Forest, and Extra Trees, as well as deep learning models such as Multi-Layer Perceptron (MLP).
\subsubsection{Joint approach}
\label{subsubsec:joint approach}
We explore the synergy between statistical and semantic approaches by conducting experiments that combine features from both methods. The goal of these experiments is to understand the complementary nature of these approaches and how their integration can enhance the accuracy of difficulty classification. Features extracted through the methods in section \ref{subsubsec:Statistical approach} and section \ref{subsubsec:Semantic approach} will be concatenated and fed into classification models, including SVM, random forest, and extra tree.
\subsubsection{Evaluation Metric}
\label{subsubsec:Evaluation Metric}

To assess the performance of the models in our experiments, we employ accuracy and F$_1$ score (macro average) as the two main evaluation metrics, where the F$_1$ score is described below:
\begin{center}
F$_1$ = $\displaystyle \frac{2\ \times \ \text{Precision}\ \times \ \text{Recall}}{\text{Precision}\ +\ \text{Recall}}$
\end{center}

\section{Experiment Result}
\label{sec:Experiment Result}
\subsection{Statistical Result}
\label{subsec:Statistical Result}
\begin{table*}[htb]
\centering
\begin{tabular}{clcc}
\hline
\multirow{2}{*}{\textbf{Dataset}}                    & \multirow{2}{*}{\textbf{Model}} & \multicolumn{2}{c}{\textbf{Result}} \\ \cmidrule{3-4} 
                                            &               & \textbf{F1} & \textbf{Acc} \\ \hline
\multicolumn{1}{c|}{\multirow{3}{*}{ViRead}} & SVM           & 88.48       & 92.05        \\
\multicolumn{1}{c|}{}                       & Random Forest & \textbf{92.59}       & \textbf{95.34}        \\
\multicolumn{1}{c|}{}                       & Extra Tree    & 91.34       & 94.52        \\ \hline
\multicolumn{1}{c|}{\multirow{3}{*}{OneStopEnglish}} & SVM                             & 72.85            & 72.81            \\ 
\multicolumn{1}{c|}{}                       & Random Forest & 74.97       & 74.56        \\
\multicolumn{1}{c|}{}                       & Extra Tree    & \textbf{75.77}       & \textbf{75.44}        \\ \hline
\multicolumn{1}{c|}{\multirow{3}{*}{RACE}}  & SVM           & 71.27       & 76.67        \\
\multicolumn{1}{c|}{}                       & Random Forest & 72.77       & 77.07        \\
\multicolumn{1}{c|}{}                       & Extra Tree    & \textbf{72.84}       & \textbf{77.07}        \\ \hline
\end{tabular}%

\caption{Statistical approach performance on machine learning model}
\label{tab:Statistical approach performance on machine learning model}
\end{table*}
The results presented in Table \ref{tab:Statistical approach performance on machine learning model} reveal the Extra Tree model performs exceptionally well on both the OneStopEnglish and RACE datasets. On the OneStopEnglish dataset, Extra Tree surpasses the other models, SVM and Random Forest, by 0.8\% in F$_1$-score compared to the second-best model (Random Forest) and by 2.92\% compared to SVM. In the RACE dataset, Extra Tree continues to be the top performer. However, the performance gap between Extra Tree and the other two models is negligible, with a 0.07\% difference with Random Forest and a 1.57\% difference with SVM in terms of F$_1$-score. This variation in performance between Extra Tree and the other models across the two datasets is likely due to the substantial difference in dataset sizes, with OneStopEnglish comprising only 567 samples, while RACE contains 27,933 samples.

In contrast to the cases in the RACE and OneStopEnglish datasets, on the ViRead dataset, Random Forest is the top-performing model with an F$_1$-score of 92.58\%, followed by Extra Tree with 91.34\%, and SVM with 88.48\%. The superior performance observed with the ViRead dataset can be attributed to the fact that the RACE and OneStopEnglish datasets are translations from English to Vietnamese. This translation process results in fewer features that are unique to Vietnamese compared to ViRead, which is derived from Vietnamese-language textbooks and thus retains more distinctive linguistic features inherent to Vietnamese.

\subsection{Semantic Result}
\label{subsec:Semantic Result}

\begin{table*}[htb]
\centering
\resizebox{\textwidth}{!}{%
\begin{tabular}{clcccccccc}
\toprule
\multicolumn{2}{c}{\multirow{4}{*}{\textbf{Semantic approach}}} & \multicolumn{8}{c}{\textbf{Result}} \\ \cmidrule{3-10}
\multicolumn{2}{c}{} & \multicolumn{4}{c}{\textbf{F1}} & \multicolumn{4}{c}{\textbf{Acc}} \\ \cmidrule{3-10}
\multicolumn{2}{c}{} & \textbf{MLP} & \textbf{SVM} & \textbf{Random Forest} & \textbf{Extra Tree} & \textbf{MLP} & \textbf{SVM} & \textbf{Random Forest} & \textbf{Extra Tree} \\ \hline
\multicolumn{1}{c|}{\multirow{3}{*}{ViRead}} & PhoBERT & 72.45 & 64.43 & 79.17 & 77.4 & 80 & 80.55 & 83.56 & 84.66 \\
\multicolumn{1}{c|}{} & ViDeBERTa & 44.45 & 14.84 & 76.34 & \textbf{80.11} & 59.73 & 42.19 & 81.92 & \textbf{84.93} \\
\multicolumn{1}{c|}{} & ViBERT & 63.17 & 62.08 & 75.36 & 73.7 & 73.7 & 77.81 & 82.19 & 83.01 \\ \hline
\multicolumn{1}{c|}{\multirow{3}{*}{OneStopEnglish}} & PhoBERT & \textbf{63.66} & 41 & 29.37 & 15.59 & \textbf{64.91} & 48.25 & 28.95 & 14.91 \\
\multicolumn{1}{c|}{} & ViDeBERTa & 40.13 & 18.56 & 55.35 & 52.32 & 46.49 & 30.7 & 54.39 & 53.51 \\
\multicolumn{1}{c|}{} & ViBERT & 41.45 & 31.02 & 32.78 & 19.66 & 42.98 & 37.72 & 33.33 & 20.18 \\ \hline
\multicolumn{1}{c|}{\multirow{3}{*}{Race}} & PhoBERT & \textbf{74.5} & 72.96 & 71.82 & 70.67 & \textbf{79.2} & 77.89 & 76.64 & 76.52 \\
\multicolumn{1}{c|}{} & ViDeBERTa & 60.16 & 56.69 & 66.22 & 64.9 & 70.93 & 70.28 & 72.1 & 72.12 \\
\multicolumn{1}{c|}{} & ViBERT & 70.01 & 68.92 & 69.06 & 66.81 & 75.47 & 75.8 & 74.65 & 74.13 \\ \hline
\end{tabular}%
}
\caption{Semantic approach using both pre-trained language models and machine learning model}
\label{tab:Semantic approach using both pre-trained language models and machine learning model}
\end{table*}
The experimental results using the language representation capabilities of pre-trained language models are summarized in Table \ref{tab:Semantic approach using both pre-trained language models and machine learning model}. The statistical results demonstrate that PhoBERT's semantic representation outperforms ViDeBERTa and ViBERT on the OneStopEnglish and RACE datasets, achieving a 63.66\% F$_1$ score on the OneStopEnglish dataset and a 74.5\% F$_1$ score on the RACE dataset when using MLP for classification. However, on the OneStopEnglish dataset, when employing other classification models such as Random Forest and Extra Tree, features extracted through PhoBERT yield lower results in both F1$_1$ score and accuracy compared to features extracted through ViDeBERTa. Nevertheless, when using SVM for classification, features extracted through PhoBERT outperform those extracted through ViDeBERTa. This discrepancy may be attributed to the small training dataset size in the OneStopEnglish dataset, leading to unusual model performance variations, unlike the RACE dataset where the performance of classification models using features extracted through PhoBERT consistently outperforms those using ViDeBERTa and ViBERT.

Similarly, the performance of classification models using features extracted through PhoBERT is generally higher than ViDeBERTa, except for one exceptional case when classifying with the Extra Tree model. In this case, the ViDeBERTa embeddings outperform PhoBERT embeddings by 2.71\% in terms of F$_1$ score and 0.27\% accuracy. This anomaly may be attributed to the small dataset size, leading to unclear and unstable differences between the two embedding methods.

Furthermore, significant variations in results are observed when comparing the performance of models determining difficulty through the semantic representation of pre-trained language models with conventional classification models using features derived from statistics. For instance, on the ViRead and OneStopEnglish datasets, the models with combined semantic and statistical features yield lower results than those employing only statistical features. This could be attributed to the limited size of the training data, causing a decrease in performance, contrary to the models trained on the RACE dataset. However, the RACE dataset needs more Vietnamese language features, resulting in only marginal performance improvement.

\subsection{Joint Result}
\label{subsec:Joint Result}

\begin{table*}[]
\centering
\resizebox{\textwidth}{!}{%
\begin{tabular}{c|lcccccccc}
\toprule
\multicolumn{2}{c}{\multirow{4}{*}{\textbf{Joint Approach}}} & \multicolumn{8}{c}{\textbf{Result}} \\\cmidrule{3-10}
\multicolumn{2}{c}{} & \multicolumn{4}{c}{\textbf{F1}} & \multicolumn{4}{c}{\textbf{Acc}} \\ \cmidrule{3-10}
\multicolumn{2}{c}{} & MLP & SVM & Random Forest & Extra Tree & MLP & SVM & Random Forest & Extra Tree \\ \hline
\multirow{3}{*}{ViRead} & PhoBERT & 91.76 & 87.52 & \textbf{92.17} & 90.06 & 94.52 & 92.05 & \textbf{94.52} & 93.15 \\
 & ViDeBERTa & 91.23 & 87.84 & 91.92 & 92.15 & 94.25 & 91.33 & 94.25 & 94.52 \\ 
 & ViBERT & 86.2 & 86.37 & 90.82 & 89.35 & 91.51 & 90.11 & 93.7 & 92.33 \\ \hline
\multirow{3}{*}{OneStopEnglish} & PhoBERT & 67.96 & 72.66 & 56.26 & 45.2 & 69.3 & 73.68 & 56.14 & 45.61 \\
 & ViDeBERTa & 67.29 & \textbf{73.72} & 64.91 & 64.51 & 70.18 & \textbf{73.88} & 64.35 & 64.91 \\
 & ViBERT & 56.33 & 71.55 & 60.93 & 49.54 & 58.77 & 72.64 & 61.4 & 50 \\ \hline
\multirow{3}{*}{Race} & PhoBERT & 73.17 & 71.62 & 73.97 & \textbf{77.09} & 78.27 & 77.69 & \textbf{78} & 77.2 \\ 
 & ViDeBERTa & 64.34 & 70.98 & 73.02 & 69.85 & 74.53 & 76.53 & 77.33 & 75.2 \\ \
 & ViBERT & 71.27 & 71.19 & 72.46 & 71.07 & 77.6 & 76.67 & 76.67 & 76.43\\ \hline
\end{tabular}%
}
\caption{Joint approach result when combine both statistical and embedding features}
\label{tab:Joint approach result when combine both statistical and embedding features}
\end{table*}
The experimental results of the classification models with the combination of features, including embeddings from pre-trained language models and statistical features, are summarized in Table \ref{tab:Joint approach result when combine both statistical and embedding features}. Overall, across the three datasets, the feature combination method significantly improves the performance of the models compared to using only features extracted by transformers (see section \ref{subsec:Semantic Result}). 

In the ViRead and OneStopEnglish datasets, the classification models' performance increases from 17.255\% to over 37.01\% in terms of F$_1$ score and from 11.3675\% to 27.41\% in terms of accuracy across the three different feature extraction methods. However, in the RACE dataset, the performance improvement of the models is not substantial, only increasing by an average of 4\% across all three embedding methods. Additionally, some cases show that the model's performance decreases when combining features, such as SVM and MLP, when extracted by PhoBERT. This is likely because the SVM and MLP models rely on certain Vietnamese-specific features that are less present in the RACE dataset than in the ViRead dataset.

Although the combined feature results are slightly lower than using only statistical features (see section \ref{subsec:Statistical Result})—lower by 0.42\% in F$_1$ score and 0.82\% in accuracy on the ViRead dataset, and 2.05\% in F$_1$ score and 1.56\% in accuracy on the OneStopEnglish dataset—the small size of these two datasets may contribute to this observation. If the dataset size is increased, as in the case of the RACE dataset, where combining features improves performance, then combining features is likely to lead to improvements in readability classification.

\section{Experiment Analysis}
\label{sec:Experiment analysis}

We utilized the best-performing models on each dataset from Section \ref{subsec:Joint Result} and further conducted individual experiments on each group of features, including statistical features and features obtained through pre-trained language models. The experimental results are summarized in Table \ref{tab:features-effect}.

Generally, the feature group that most influence the models when combining statistical and embedding features is the 'Raw Feature',' followed by 'POS Feature,' 'Word Cohesion', 'Syntax-Level Feature,' and finally the 'Vietnamese-Specific Feature'. The improvement in model performance when using the 'Raw Feature' group alone is understandable. This is because texts with many sentences and words per sentence encompass vast knowledge, directly influencing the text's difficulty by requiring readers to absorb a significant amount of information. Combining features from the 'Raw Feature' group with machine learning models significantly enhances the model's performance.

Apart from the 'Raw Feature' group, the 'POS Feature' and 'Word Cohesion' feature groups also affect the model's performance. In 'POS Feature,' if a text contains many polysemous words, the complexity of the text increases, requiring readers to understand the context of the sentence to truly comprehend the intended meaning of the ambiguous word. In the 'Word Cohesion' group, features representing the relationships between words and sentences within a paragraph increase the text's difficulty, demanding that readers link information within the same sentence and paragraph to form a complete data set.

While not significantly improving the model's performance like the three feature groups mentioned above, the' Syntax-Level Features' group still contributes to determining the sentence's difficulty through conjunction words. If the number of conjunction words is high, it creates multiple layers of relationships between subjects, a phenomenon present in the sentence. In contrast to the other feature groups, the 'Vietnamese-Specific Feature' group decreases the models' performance on all three datasets. This may be because the statistical features we used do not accurately reflect the nature of specific features present in Vietnamese. Sino-Vietnamese and borrowed words may indicate different semantic layers depending on usage, context, and the reader's existing knowledge. Therefore, determining the features of Sino-Vietnamese and borrowed words through a statistical approach may not be suitable.

Table \ref{tab:accuracy of models according to data size} from the paper provides a comparative analysis of the accuracies achieved by different machine learning models across three datasets—Luong, OneStopEnglish, and RACE—with varying amounts of data (25\%, 50\%, and 75\%). For the Luong dataset, the PhoBERT + MLP model shows a significant improvement in accuracy as the data size increases, while Random Forest and PhoBERT + Random Forest demonstrate remarkably high accuracy across all data sizes. In the case of OneStopEnglish, PhoBERT + MLP show increased accuracy with more data, but the performance is notably lower than on the Luong dataset, with PhoBERT + SVM even decreasing in accuracy as more data is provided. This could be explained that the OneStopEnglish dataset has only 567 samples, Extra Trees—a model that can capture complex patterns—might be overfitting to the training data at smaller data sizes. For the RACE dataset, the models exhibit a general trend of decreased accuracy a bit with increased data, with PhoBERT + Extra Trees showing the least variation. This may be due to the translation come with noise when increasing the size of data that can affect the model's ability to make accurate predictions. These findings underscore the importance of considering both the nature of the dataset and the volume of data when selecting models for text readability tasks. It appears that no single model consistently outperforms others across all datasets and data sizes, highlighting the necessity for tailored approaches in readability assessment.

\section{Conclusion}
In this paper, we propose a novel approach to the Vietnamese readability task by incorporating semantic features alongside traditional statistical features, leading to promising results on readability datasets. Additionally, we examine the impact of combining both feature types to enhance the performance of existing models. Our research has the potential to support the development of readability assessment systems for elementary-level writing. Using our model, educators can gain clear insights into the strengths and limitations of young students' essays, thereby aiding the learning and writing process in early education. Beyond this, our research shows promise in developing systems that suggest quality improvements for essays or even detect essays generated automatically by large language models.

\section*{Limitation}
While this study marks a significant advancement in the assessment of Vietnamese text readability, there are several limitations that must be acknowledged. Firstly, the reliance on translated datasets from English (OneStopEnglish and RACE) may not fully capture the intrinsic linguistic and cultural nuances of Vietnamese, potentially affecting the generalizability of the findings. Another limitation is the scope of the datasets used. The Vietnamese Text Readability Dataset (ViRead) is robust but may not represent all genres and styles of Vietnamese text. This could limit the model's applicability to diverse types of Vietnamese writings. Moreover, the machine learning models employed, despite their efficacy, might still have inherent biases and limitations in understanding complex language structures and idiomatic expressions. Finally, the current study focuses on lexical and syntactic features without deeply exploring pragmatic and discourse-level features, which are crucial for comprehensive readability assessment. 

These limitations highlight areas for future research, suggesting the need for more diverse and culturally rich Vietnamese datasets, exploration of additional language models, and a broader consideration of linguistic features for a more nuanced understanding of text readability in Vietnamese.

\bibliography{custom}

\begin{table*}[t]
\small
\begin{threeparttable}
\centering
\begin{tabular}{clcccccccccc}
\hline
\multirow{2}{*}{\textbf{Dataset}} & \multicolumn{1}{c}{\multirow{2}{*}{\textbf{Model}}} & \multicolumn{2}{c}{\textbf{Raw}} & \multicolumn{2}{c}{\textbf{POS}} & \multicolumn{2}{c}{\textbf{Syntax-Level}} & \multicolumn{2}{c}{\textbf{Viet-Spec}} & \multicolumn{2}{c}{\textbf{Word Coh.}} \\ 
\cline{3-12}
 & \multicolumn{1}{c}{} & \textbf{Acc} & \textbf{F1} & \textbf{Acc} & \textbf{F1} & \textbf{Acc} & \textbf{F1} & \textbf{Acc} & \textbf{F1} & \textbf{Acc} & \textbf{F1} \\ 
\hline
\multirow{2}{*}{ViRead} & PhoBERT + MLP & 94.79 & 92.1 & \textbf{95.07} & \textbf{92.83} & 93.7 & 91.38 & 80 & 76.84 & 93.42 & 91 \\
 & PhoBERT + RF & \textbf{93.7} & \textbf{90.7} & 92.6 & 89.83 & 90.68 & 86.69 & 83.56 & 77 & 87.67 & 82.06 \\ 
\hline
\multirow{2}{*}{OneStopEnglish} & PhoBERT + MLP & 56.14 & 46.06 & 56.14 & 55.03 & 44.74 & 36.8 & 57.02 & 54.76 & \textbf{79.09} & \textbf{70.18} \\
 & PhoBERT + SVM & \textbf{72.81} & \textbf{72.78} & 64.91 & 64.93 & 43.86 & 37.38 & 54.39 & 53.99 & 58.77 & 59.35 \\ 
\hline
\multirow{2}{*}{RACE} & PhoBERT + MLP & \textbf{78.75} & \textbf{75.35} & 78.55 & 74.46 & 78.89 & 75.34 & 77.79 & 74.11 & 78.61 & 75.24 \\
 & PhoBERT + ET & 77.63 & 72.36 & 76.78 & 71.01 & 76.96 & 71.21 & 76.56 & 70.63 & 76.7 & 70.86 \\ 
\hline
\end{tabular}
\caption[\textwidth]{The effect of statistical features on the performance of the model when combining both Embedding and statistical features}
\label{tab:features-effect}
\end{threeparttable}
\end{table*}

\begin{table*}[htbp]
\centering
\small
\begin{tabular}{clccc}
\hline
\multirow{2}{*}{Dataset} & \multicolumn{1}{c}{\multirow{2}{*}{Model}} & Acc & Acc & Acc \\
& \multicolumn{1}{c}{} & 25\% & 50\% & 75\% \\ \hline
\multirow{3}{*}{ViRead} & PhoBERT + MLP & 82.61 & 95.63 & 96.35 \\
& Random Forest & 98.91 & 99.45 & 98.18 \\
& PhoBERT + Random Forest & 92.39 & 97.81 & 97.45 \\ \hline
\multirow{3}{*}{OneStopEnglish} & PhoBERT + MLP & 37.93 & 54.39 & 65.88 \\
& Extra Trees & 86.21 & 75.44 & 80 \\
& PhoBERT + SVM & 86.21 & 68.42 & 57.65 \\ \hline
\multirow{3}{*}{RACE} & PhoBERT + MLP & 80.86 & 79.04 & 79.52 \\
& Extra Trees & 80.24 & 77.33 & 78.54 \\
& PhoBERT + Extra Tree & 78.8 & 77.65 & 77.77 \\ \hline
\end{tabular}
\caption{Accuracy of models according to data size}
\label{tab:accuracy of models according to data size}
\end{table*}
\pagebreak
\appendix

\section{Analysis of different features on the performance}

In Table \ref{tab:features-effect}, we present the impact of different feature groups on the performance of models that combine both embedding and statistical features. These experiments were conducted using the best-performing models from each dataset. The results demonstrate that the ``Raw Feature'' group has the most significant effect on model performance, followed by the ``POS Feature'' and ``Word Cohesion'' groups. In contrast, ``Syntax-Level'' and ``Vietnamese-Specific'' features contribute less to performance improvement, with Vietnamese-specific features sometimes leading to decreased performance compared to raw features.

\section{Analysis of performance based on the data size}

Table 8 presents a comparison of model accuracies across datasets with varying data sizes (25\%, 50\%, and 75\%). The results demonstrate how accuracy trends vary depending on the dataset and the model used. While PhoBERT-based models like PhoBERT + MLP show consistent improvement with larger data sizes in most cases, others like Random Forest exhibit stable high accuracy across all data sizes.

\end{document}